\documentclass[letterpaper, 10 pt, conference]{ieeeconf_marginfix}  % Comment this line out if you need a4paper

\IEEEoverridecommandlockouts                              % This command is only needed if 
\usepackage{graphics} % for pdf, bitmapped graphics files
\usepackage{times} % assumes new font selection scheme installed
\usepackage{amsmath} % assumes amsmath package installed
\usepackage{amssymb}  % assumes amsmath package installed

\usepackage{balance}
\usepackage[svgnames]{xcolor}
\usepackage{colortbl}
\usepackage{url}
\usepackage{svg}
\usepackage{array}
\usepackage{multirow}
\usepackage{makecell}       % for adding linebreaks within tables
\usepackage{cite}
\usepackage{arydshln}
\usepackage{nicematrix}
\usepackage{subcaption}
\usepackage{graphbox}
\usepackage{soul}
\usepackage{tabularx}
\usepackage{threeparttable}
\usepackage{siunitx}
\usepackage{algorithm}
\usepackage{algorithmic}
\usepackage{threeparttable}
\usepackage{booktabs}
\usepackage{hyperref}
\hypersetup{
    colorlinks=true,
    linkcolor=blue,
    filecolor=magenta,
    citecolor=blue,
    urlcolor=cyan,
}
\usepackage{pifont}% http://ctan.org/pkg/pifont
\usepackage[accsupp]{axessibility}  % Improves PDF readability for those with disabilities.
\usepackage[capitalize]{cleveref} % load this last
\crefname{section}{Sec.}{Secs.}
\Crefname{section}{Section}{Sections}
\crefname{table}{Tab.}{Tabs.}
\Crefname{table}{Table}{Tables}

\newcommand{\eg}{\emph{e.g.},}

\DeclareSIUnit{\pp}{p.p.}
\DeclareSIUnit{\nothing}{\relax}

\makeatletter
\let\old@makecaption\@makecaption
\renewcommand{\@makecaption}[2]{\old@makecaption{\small #1}{\small #2}}
\makeatother

\title{\LARGE \bf
Stereo Visual Odometry with In-the-Loop Geometric Feed-Forward Models Towards Dense Metric Mapping
}

\author{Fahira Afzal Maken$^1$, Sagun Man Singh Shrestha$^1$, Edward Vanderfeen$^{1,2}$, Lucas Carvalho de Lima$^1$,\\ Rongxin Li$^1$, Changming Sun$^1$ and Milad Ramezani$^{1,2*}$
\thanks{
$^1$ CSIRO Robotics, Technology, CSIRO, Brisbane, Australia. 
E-mails: {\tt\footnotesize \emph{
firstname.lastname
}@csiro.au}}
\thanks{
$^{2}$ School of Electrical Engineering and Robotics, Queensland University of Technology (QUT), Brisbane, Australia.}
\thanks{
$^*$Corresponding author
}
}

\begin{document}

\maketitle

\begin{abstract}
We present a stereo visual odometry (SVO) framework that integrates geometric feed-forward models (GFMs) for dense depth prediction with factor-graph-based visual odometry to achieve accurate localisation and dense metric mapping. Our key idea is to exploit sparse but reliable metric geometry estimated from stereo and temporal multi-view observations to calibrate dense GFM predictions, while retaining the robustness and efficiency of model-based SVO. Rather than directly conditioning the GFM on potentially noisy stereo depth, optimised $3$D landmarks are used to jointly estimate a scale--shift transformation over all keyframes within each local submap, yielding metrically consistent dense depth predictions. To balance accuracy and computational efficiency, we introduce a hierarchical factor-graph architecture. At the local level, sliding-window optimisation estimates high-frequency camera poses and multi-view-refined $3$D landmarks. Dense predictions are generated only for selected keyframes and registered within each submap using $\mathrm{SE}(3)$ Generalised ICP. Consecutive submaps are further aligned in $\mathrm{Sim}(3)$ using their shared geometry, allowing residual scale inconsistencies to be corrected. These intra- and inter-submap constraints are incorporated into a global factor graph to produce a consistent trajectory and dense metric reconstruction.
Extensive experiments on public and in-house datasets demonstrate that the proposed framework achieves competitive odometry accuracy against geometric and learning-based baselines while providing accurate dense metric reconstruction and improved robustness to failures in learned depth prediction.
\end{abstract}

\section{Introduction}
\label{sec:introduction}

% \begin{figure}[t]
%     \centering
%     \includegraphics[width=0.99\linewidth]{figures/hero.jpg}
%     \caption{\sagun{Top: the stereo camera and the lidar on the in-house rig; the lidar serves only as the reference. Middle: stereo feature tracking in the left and right images. Bottom left: the dense metric map of the in-house sequence, submaps in distinct colours over the reference lidar map (grey), with the estimated (red) and reference (cyan) trajectories; the loop closes on its start (magenta) without any revisit detection. Bottom right: the RGB map seen from the camera on the return leg, looking along the corridor towards the start (green) and end (magenta) of the loop.}}
%     \label{fig:hero}
% \end{figure}

% NOTE: alternative hero figure (both bottom panels from the same first-person view); keep one of the two and comment the other out.
\begin{figure}[t]
    \centering
    \includegraphics[width=0.99\linewidth]{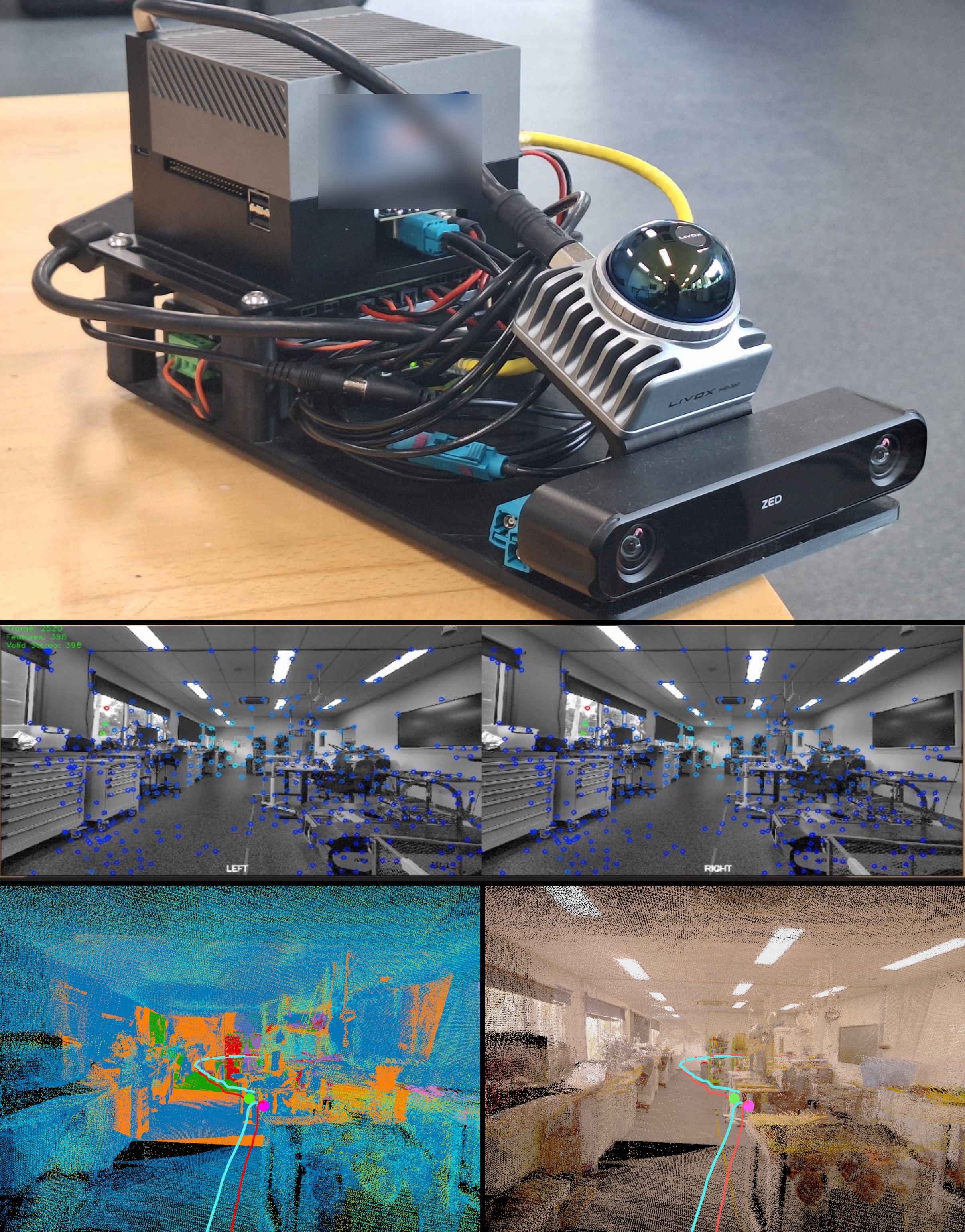}
    \caption{\textbf{Top}: stereo camera and LiDAR mounted on the in-house rig; the LiDAR is used only as reference. \textbf{Middle}: stereo feature tracking in the left and right images. \textbf{Bottom}: dense metric reconstruction of the in-house sequence, shown with submaps in distinct colours (left) and in RGB (right), viewed from the return leg along the corridor towards the start (green) and end (magenta) of the loop, together with the estimated (red) and reference (cyan) trajectories.}
    \label{fig:hero}
    \vspace{-5mm}
\end{figure}

Recent advances in deep learning have transformed visual simultaneous localisation and mapping (SLAM), with geometric feed-forward models (GFMs) showing strong capability in recovering dense $3$D scene structure directly from images. Building on these advances, systems such as MASt3R-SLAM~\cite{11094566} and VGGT-SLAM~\cite{maggio2026vggt} have demonstrated impressive localisation and reconstruction performance. However, dense inference remains computationally demanding, inconsistencies across local reconstructions must be reconciled over long trajectories, and predicted depth may still exhibit residual scale inconsistencies with metric geometry.

While recent feed-forward SLAM systems can recover dense scene geometry, maintaining globally consistent and metrically scaled reconstructions over long trajectories remains challenging. Approaches such as VGGT-SLAM address reconstruction ambiguity through global alignment and optimisation of local submaps~\cite{NEURIPS2025_bc65ab11}, although over the $\mathrm{SL}(4)$ manifold with 15 degrees of freedom (DoF). On the other hand, feature-based stereo visual odometry (SVO)~\cite{murORB2,9440682,mei2011rslam} provides accurate metric geometry through joint optimisation of camera poses and visual landmarks, leveraging both stereo and temporal multi-view observations. However, conventional SVO typically produces only sparse scene structure. This motivates combining the metric consistency of stereo geometry with the dense reconstruction capability of GFMs.

In this work, we present an SVO framework that integrates geometric feed-forward depth prediction with factor-graph optimisation through a hierarchical architecture. At the local level, sliding-window factor graphs estimate high-frequency camera poses and multi-view-refined $3$D landmarks. For each local submap, the optimised landmarks from all selected keyframes are jointly used to estimate a scale--shift transformation that metrically calibrates the dense GFM predictions. The resulting dense reconstructions are registered between keyframes, while consecutive local submaps are aligned through $\mathrm{Sim}(3)$ transformations to correct residual scale discrepancies. These geometric constraints are incorporated into a global factor graph to provide accurate localisation and globally consistent dense metric mapping (\Cref{fig:hero}). By restricting dense inference and registration to selected keyframes, the proposed hierarchy balances reconstruction accuracy with computational efficiency.

The main contributions of this work are:
\begin{itemize}

\item We propose a hierarchical factor-graph stereo visual odometry framework that combines sparse metric geometry with dense geometric feed-forward predictions, using $\mathrm{SE}(3)$ constraints within local submaps and $\mathrm{Sim}(3)$ alignment across submaps for globally consistent dense metric mapping.

\item We introduce a metric-guided depth calibration strategy that uses multi-view-optimised $3$D landmarks to estimate a shared scale--shift transformation for all keyframes within each submap, providing metrically consistent dense depth predictions without directly conditioning the GFM on uncertain stereo depth.

\item We extensively evaluate the proposed framework on KITTI, Newer College, and in-house datasets against representative geometric and learning-based baselines, considering both trajectory accuracy and dense reconstruction quality.

\end{itemize}
\section{Related Works}
\label{sec:relatedworks}
We focus on visual-only odometry and SLAM, and therefore exclude approaches that rely on additional sensing modalities such as inertial measurements. We review classical visual odometry, recent feed-forward geometric models, and learning-based SLAM systems relevant to our approach.

\subsection{Visual Odometry}
Visual odometry (VO) estimates camera motion from consecutive images using geometric constraints derived from visual observations. Early stereo VO demonstrated the benefits of directly measuring metric scale for applications ranging from planetary exploration~\cite{goldberg2002stereo} to autonomous ground vehicles~\cite{4651147}. Modern feature-based systems typically detect and associate sparse visual features, initialise their $3$D structure from stereo observations, and jointly refine camera poses and landmarks using multi-view geometric optimisation. ORB-SLAM2~\cite{murORB2} extended feature-based SLAM to monocular, stereo, and RGB-D cameras, while ORB-SLAM3~\cite{9440682} further improved multi-map and long-term operation. Related graph-based systems such as RSLAM~\cite{mei2011rslam} and OpenVSLAM~\cite{10.1145/3343031.3350539} similarly exploit sparse landmark observations to maintain geometrically consistent camera trajectories. These approaches provide accurate and computationally efficient metric estimation when stereo observations are available, but their maps are inherently sparse and capture only a small fraction of the observed scene geometry.

\subsection{Deep Scene Reconstruction}
Recent vision foundation models have substantially advanced image-based depth and $3$D reconstruction. Transformer-based models such as Depth Anything~\cite{lin2026depth} learn strong monocular and multi-view depth priors, while MVSAnywhere~\cite{izquierdo2025mvsanywhere} combines monocular and multi-view cues for generalisable multi-view stereo. More general models such as MapAnything~\cite{11533250} accept images together with optional geometric information, including camera parameters, depth, and partial reconstructions, to predict metric scene geometry.

A parallel line of work directly reasons about $3$D structure across images. DUSt3R~\cite{dust3r_cvpr24} predicts dense pointmaps from pairs of overlapping images without requiring known camera calibration, while MASt3R~\cite{mast3r_eccv24} extends this formulation with improved geometric matching. These models provide powerful dense reconstruction priors without requiring conventional feature extraction and triangulation. However, applying pairwise or window-based predictions over long sequences requires the resulting local reconstructions to be aligned and reconciled globally. Related depth-completion methods also demonstrate the benefit of combining learned priors with sparse metric observations. For example, Marigold-DC~\cite{11444687} guides an affine-invariant monocular depth prior using sparse metric depth measurements, although its iterative diffusion-based inference is not designed as a feed-forward VO framework. These works motivate exploiting sparse geometric measurements to metrically constrain learned dense predictions.

\subsection{Deep Visual SLAM}
Deep learning has also been integrated directly into visual SLAM. 
DROID-SLAM~\cite{NEURIPS2021_89fcd07f} jointly updates pixel-wise depth and 
camera poses through a recurrent neural network and differentiable dense 
bundle adjustment, and can exploit monocular, stereo, or RGB-D imagery. 
More recently, geometric foundation models have been incorporated into hybrid 
SLAM pipelines. MASt3R-SLAM~\cite{11094566} builds a real-time dense monocular 
SLAM system around MASt3R, combining learned pointmaps with camera tracking, 
local fusion, loop closure, and global graph optimisation.

VGGT-SLAM~\cite{NEURIPS2025_bc65ab11} instead forms local dense submaps using 
feed-forward VGGT~\cite{wang2025vggt} predictions and globally aligns them through factor-graph 
optimisation. To account for reconstruction ambiguity with uncalibrated 
monocular imagery, it estimates 15-DoF projective transformations between 
submaps on the $\mathrm{SL}(4)$ manifold. VGGT-SLAM 2.0~\cite{maggio2026vggt} 
improves this formulation through a keyframe-level factor graph that constrains 
the high-dimensional projective ambiguity and reduces the drift and degeneracy 
associated with unconstrained 15-DoF submap alignment.

\begin{figure*}[t]
    \centering
    \includegraphics[width=0.95\linewidth]{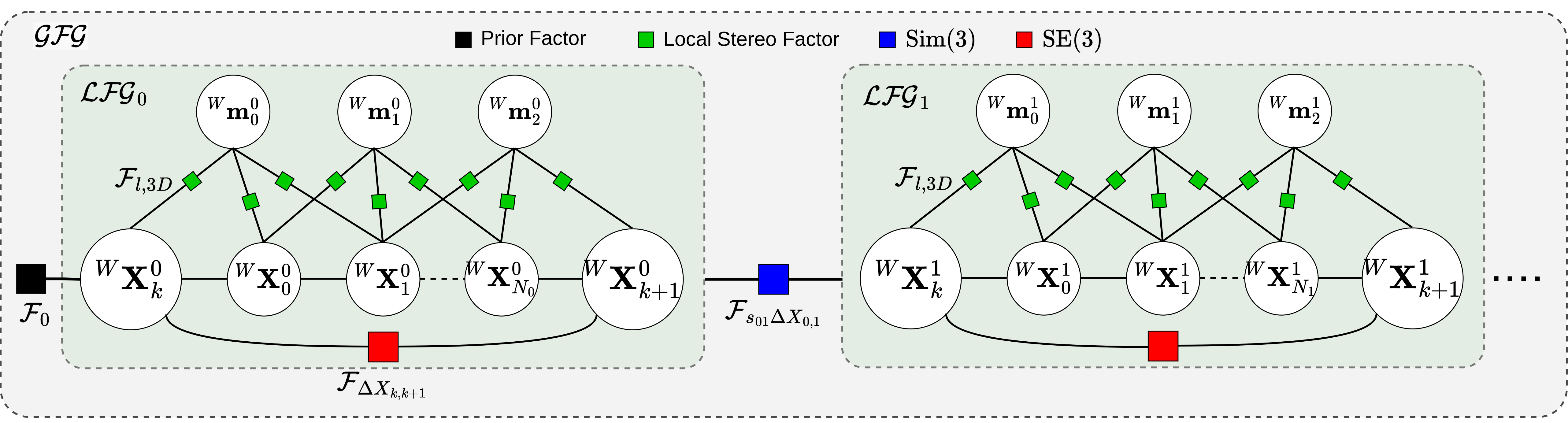}
    \caption{Factor structure of our SVO. Each LFG jointly optimises camera poses and $3$D landmarks using local stereo factors, while  $SE(3)$ ICP factors connect keyframes within the window of LFGs and $Sim(3)$ factors across consecutive LFGs. The GFG maintains the overall trajectory using prior and inter-keyframe as well as between local window constraints. For simplicity, the superscript indicating the LFG count is omitted from $\mathbf{m}$ and $\mathbf{X}$ throughout the text.}
    \label{fig:pipeline}
    % Subcaptions
    \vspace{-4mm}
\end{figure*}

ScaRF-SLAM~\cite{zhang2026scarf} decouples feature-based visual SLAM from GFM-based dense reconstruction, using the former for tracking and the latter for mapping, with frame- and submap-level scale optimisation for consistency. While this prevents GFM errors from directly affecting pose estimation, the dense reconstruction is not used to further improve localisation. Moreover, these GFM-based approaches do not inherently recover metric scale.

% Our approach follows a similar principle of preserving model-based visual 
% odometry for tracking, but differs in how metric geometry is incorporated into 
% the dense reconstruction. Rather than relying primarily on camera poses and 
% depth-scale optimisation, we exploit stereo and temporal multi-view geometry to 
% estimate metric 3D landmarks within local factor graphs. These optimised 
% landmarks directly establish sparse metric depth correspondences for affine 
% calibration of the GFM predictions, while consecutive local reconstructions are 
% aligned through $\mathrm{Sim}(3)$ transformations.

\section{Method} \label{sec:method}
\subsection{System Overview}
\label{sec:overview}
Given a sequence of synchronised and calibrated stereo image pairs
$\mathcal{I}=\{I_f^l,I_f^r\}_{f=0}^{N_f}$, our goal is to estimate the camera trajectory while constructing a globally consistent dense metric reconstruction of the environment. As illustrated in~\Cref{fig:pipeline}, the proposed framework follows a hierarchical factor-graph architecture composed of a set of local factor graphs $\{\mathcal{LFG}_i\}_{i=0}^{N_L}$ for high-frequency visual odometry and a global factor graph ($\mathcal{GFG}$) to maintain long-term global trajectory and dense-map consistency.

At the local level, stereo and temporal multi-view observations are jointly
optimised within a sliding window to estimate camera poses
$\mathbf{X}_f\in\mathrm{SE}(3)$ and $3$D visual landmarks
$\mathbf{m}_j\in\mathbb{R}^3$. Stereo observations provide metric scale
through the known camera baseline, while temporal observations across multiple
frames improve landmark geometry and pose estimation. Each $\mathcal{LFG}_i$
therefore provides a sparse and geometrically consistent local reconstruction.

For selected keyframes, the optimised $3$D landmarks provide sparse metric depth
constraints for metrically aligning the dense depth predicted by a geometric
feed-forward model (GFM). The resulting metrically aligned depth maps are
back-projected into dense point clouds and registered across keyframes to form
submaps. We align consecutive submaps through a similarity transformation in
$\mathrm{Sim}(3)$ using their overlapping geometry, allowing residual
rotation, translation, and scale inconsistencies between independently
optimised local graphs to be corrected.

At the global level, the registration constraints are combined with the relative motion estimates inherited from the local SVO within the $\mathcal{GFG}$. Optimising this graph provides globally consistent keyframe poses and aligns the dense keyframe reconstructions within a common metric reference frame. By performing local geometric optimisation at every frame while invoking dense feed-forward inference and global registration only at selected keyframes, the proposed hierarchy aims to preserve the geometric accuracy of multi-view stereo while reducing the computational cost of dense reconstruction.

\subsection{Local Stereo Visual Odometry}
\label{sec:local_svo}

Each local factor graph $\mathcal{LFG}_i$ operates over a sliding window containing a variable number of frames and keyframes, where keyframes are selected by camera motion and scene covisibility. Within each $\mathcal{LFG}$, camera poses $\mathbf{X}_f\in\mathrm{SE}(3)$ and $3$D visual landmarks $\mathbf{m}_j\in\mathbb{R}^3$ are jointly optimised using stereo and temporal multi-view observations. Features are tracked temporally in the left-image sequence using the KLT tracker~\cite{tomasi1991detection}, while stereo correspondences are established between synchronised left and right images using window-based matching~\cite{hirschmuller2005accurate}.
% ~\milad{ref. I know we use KLT here too but perhaps not good to mention KLT here}. 
% A stereo observation is represented as
\begin{equation}
    \mathbf{z}_{fj}
    =
    \begin{bmatrix}
        u^{l}_{fj} & u^{r}_{fj} & v_{fj}
    \end{bmatrix}^{\top}
    \in\mathbb{R}^{3},
\end{equation}
where \(u^{l}_{fj}\) and \(u^{r}_{fj}\) are the horizontal image coordinates of feature \(j\) in the left and right images, respectively, and \(v_{fj}\) is its vertical image coordinate.

Camera poses are initialised using a Perspective-$n$-Point (PnP) solver~\cite{8099974}, followed by sliding-window factor-graph optimisation over stereo and temporal multi-view observations. Let \(\mathcal{O}_i\) denote the feature-observation set associated with local factor graph \(i\).
Assuming zero-mean Gaussian measurement noise with covariance $\boldsymbol{\Sigma}_{fj}$, the local estimate is obtained by minimising the stereo reprojection error,
\begin{equation}
    \underset{\{\mathbf{X}_f,\mathbf{m}_j\}}{\text{minimise}}
    \sum_{(f,j)\in\mathcal{O}_i}
        \left\|
            \mathbf{z}_{fj}
            -
            \mathbf{h}(\mathbf{X}_f,\mathbf{m}_j)
        \right\|_{\boldsymbol{\Sigma}_{fj}}^{2},
\end{equation}
where $\mathbf{h}(\cdot)$ denotes the calibrated stereo projection model.

The optimisation jointly estimates metrically scaled camera poses and multi-view-refined landmarks. For each selected keyframe, these landmarks are used to provide sparse metric depth guidance to the geometric feed-forward model.

\subsection{Metric-Guided Dense Depth Prediction}
\label{sec:metric_depth}

For each selected keyframe $k$, the GFM predicts a dense depth map
$\hat{\mathbf{D}}_k$. Although recent GFMs can recover dense scene structure,
their predictions may still exhibit residual scale-shift and inconsistencies
with respect to metric geometry. Rather than directly conditioning the GFM on sparse
metric measurements, such as raw stereo depth, which becomes increasingly
imprecise at long range, we retain the model's original inference process and
metrically align its output using the optimised $3$D landmarks from the
corresponding LFG. Unlike raw stereo measurements, these landmarks are
jointly refined through stereo and temporal multi-view geometry, providing a more
reliable sparse metric local scene representation for aligning dense depth predictions.

Rather than independently aligning each keyframe, we estimate a single
scale--shift transformation for all keyframes within each local submap.
Let $\mathcal{K}_i$ denote the set of keyframes in $\mathcal{LFG}_i$.
Following recent scale--shift depth alignment approaches~\cite{11444687,lin2026depth}, the metric depth for each
$k\in\mathcal{K}_i$ is given by
\begin{equation}
    \mathbf{D}_k^{m}
    =
    a_i \hat{\mathbf{D}}_k + b_i ,
    \label{eq:metric_depth}
\end{equation}
where $a_i$ and $b_i$ are the scale and shift shared by the entire submap,
and $\mathbf{D}_k^{m}$ denotes the resulting metric depth map.

Let $\mathcal{A}_k$ denote the set of optimised landmarks visible in keyframe
$k$. Each landmark $\mathbf{m}_j \in \mathcal{A}_k$ is transformed into the coordinate frame of
the corresponding keyframe using the optimised pose of keyframe $k$ and projected onto the image, yielding a sparse
metric depth $d_{kj}$ at pixel $\mathbf{u}_{kj}$. The GFM prediction at the
same location is
\begin{equation}
    \hat{d}_{kj}
    =
    \hat{\mathbf{D}}_k(\mathbf{u}_{kj}) .
\end{equation}

The common affine parameters for $\mathcal{LFG}_i$ are estimated jointly from
all landmark--depth correspondences across its keyframes using iteratively
reweighted least squares (IRLS),
\begin{equation}
    \underset{a_i,b_i}{\text{minimise}}
    \sum_{k\in\mathcal{K}_i}
    \sum_{j\in\mathcal{A}_k}
        w_{kj}
        \left(
            d_{kj}
            -
            a_i \hat{d}_{kj}
            -
            b_i
        \right)^2 ,
    \label{eq:depth_irls}
\end{equation}
where $w_{kj}$ is iteratively updated according to the residual to reduce the
influence of inconsistent depth correspondences. After convergence, the same
estimated scale and shift are applied to all dense predictions within the
submap using~\eqref{eq:metric_depth}, producing dense depth maps expressed in
the metric reference established by the local SVO.

% \sagun{In the experiments of~\Cref{sec:experiments} the shift is fixed, $b_i=0$, because the landmarks of an indoor submap span a narrow depth range and the fitted scale and shift correlate at approximately $-0.99$; the per-submap scales are then smoothed across consecutive submaps with a one-dimensional factor graph over their logarithms, in which each submap contributes a prior from its own fit and, when enough landmarks are shared with the previous submap, a between factor computed from those landmarks.}
% Feel free to remove

\subsection{Dense Map Registration}
\label{sec:dense_registration}
For each keyframe $k$, the metrically aligned depth map $\mathbf{D}_k^{m}$ is
back-projected using the calibrated camera model to obtain a dense point cloud
$\mathcal{P}_k$. Within each local submap, overlapping keyframes are registered
using Generalised ICP (GICP)~\cite{segal2009generalized}. Since keyframes within
the same $\mathcal{LFG}_i$ share a common metric reference, their relative
transformation is estimated in $\mathrm{SE}(3)$ as
\begin{equation}
    \mathbf{T}_{k,k+1}^{\mathrm{GICP}}
    =
    \underset{\mathbf{T}\in\mathrm{SE}(3)}{\operatorname{argmin}}
        \;
        \mathcal{J}_{\mathrm{GICP}}
        \left(
            \mathcal{P}_k,
            \mathcal{P}_{k+1},
            \mathbf{T}
        \right),
\end{equation}
where $\mathbf{T}_{k,k+1}^{\mathrm{GICP}}$ aligns $\mathcal{P}_{k+1}$ with
$\mathcal{P}_k$. These registrations provide relative geometric constraints
between keyframes within each submap.

Consecutive local submaps are constructed with at least one shared keyframe,
\eg~the last keyframe of $\mathcal{LFG}_i$ is also used as the first keyframe
of $\mathcal{LFG}_{i+1}$. Although the two submaps observe common scene
geometry, they are independently optimised and may retain a residual scale
discrepancy. The shared keyframe provides reliable $3$D correspondences between
the two local reconstructions. We first use RANSAC~\cite{10.1145/358669.358692} over these correspondences
to obtain a robust initial estimate of the relative scale $s_{i,i+1}^{0}$. Using this scale initialisation, the alignment between consecutive submaps is
then refined via GICP over the similarity group,
\vspace{-2mm}
\begin{equation}
    \mathbf{S}_{i,i+1}^{\mathrm{GICP}}
    =
    \underset{\mathbf{S}\in\mathrm{Sim}(3)}{\operatorname{argmin}}
        \;
        \mathcal{J}_{\mathrm{GICP}}
        \left(
            \mathcal{P}_{i},
            \mathcal{P}_{i+1},
            \mathbf{S}%;
           % s_{i,i+1}^{0}
        \right)%,
\text{initialized at } s^{0}_{i,i+1},
\end{equation}
where $\mathbf{S}_{i,i+1}^{\mathrm{GICP}}$ jointly refines the relative
rotation, translation, and scale between the two submaps. The resulting
$\mathrm{SE}(3)$ keyframe registrations and $\mathrm{Sim}(3)$ inter-submap
alignments are then incorporated into the global factor graph.

\subsection{Global Factor-Graph Optimisation}
\label{sec:global_fgo}
The global factor graph $\mathcal{GFG}$ combines $\mathrm{SE}(3)$ GICP
constraints between keyframes within each submap with $\mathrm{Sim}(3)$
constraints between consecutive submaps. Let
$\mathcal{X}_{G}=\{\mathbf{X}_k,\mathbf{S}_i\}$ denote the global graph
variables for  keyframe poses and submap similarity transformations.
The global estimate is obtained by jointly minimising both residuals,
\begin{equation}
    \resizebox{\linewidth}{!}{$\displaystyle
        \underset{\mathcal{X}_{G}}{\text{minimise}}
        \left[
            \sum_{(k,k+1)\in\mathcal{E}_{\mathrm{SE(3)}}}
                \left\|
                    \mathbf{r}_{k,k+1}^{\mathrm{SE(3)}}
                \right\|_{\boldsymbol{\Sigma}_{k,k+1}}^{2}
            +
            \sum_{(i,i+1)\in\mathcal{E}_{\mathrm{Sim(3)}}}
                \left\|
                    \mathbf{r}_{i,i+1}^{\mathrm{Sim(3)}}
                \right\|_{\boldsymbol{\Sigma}_{i,i+1}}^{2}
        \right],
    $}
    \label{eq:global_fgo}
\end{equation}
where $\mathcal{E}_{\mathrm{SE(3)}}$ and
$\mathcal{E}_{\mathrm{Sim(3)}}$ denote the intra- and inter-submap edges,
respectively. Covariances $\boldsymbol{\Sigma}_{k,k+1}$ and
$\boldsymbol{\Sigma}_{i,i+1}$ encode the uncertainty associated with the
corresponding $\mathrm{SE(3)}$ and $\mathrm{Sim(3)}$ registration constraints.
The optimised transformations place all dense keyframe
reconstructions in a common metric reference frame. 
% \sagun{Besides the registration constraints, the $\mathcal{GFG}$ retains the relative odometry between consecutive keyframes and the stereo factors of the landmarks used for metric alignment in~\Cref{sec:metric_depth}.}
% Feel free to remove
\section{Experiments}
\label{sec:experiments}

% We evaluate the proposed framework on the KITTI Odometry benchmark~\cite{geiger2012we}, the Newer College Dataset (NCD)~\cite{ramezani2020newer}, and an in-house stereo dataset. The experiments are designed to assess: 1) trajectory accuracy over large-scale and challenging sequences; 2) the quality and global consistency of the reconstructed dense metric maps; and 3) the computational cost of the proposed hierarchical architecture.

\subsection{Implementation Details}
\label{sec:implementation}
Our system is implemented using GTSAM~\cite{dellaert2012factor} for
factor-graph-based stereo visual odometry and MapAnything~\cite{11533250} as
the GFM for dense depth prediction. %Local factor graphs are optimised within a sliding window using iSAM2~\cite{5979641}. 
In the front-end, we use GFTT~\cite{323794} for feature detection and track them temporally using KLT~\cite{tomasi1991detection} as described earlier.  As new frames arrive, the local factor graph is incrementally updated using iSAM2~\cite{5979641}. 
% For each local submap, dense depth inference is performed only on selected
% keyframes, and a single scale--shift transformation is estimated jointly from
% the optimised 3D landmarks across all keyframes in the submap. Dense
% reconstructions are registered between keyframes in $\mathrm{SE}(3)$, while
% consecutive submaps are aligned in $\mathrm{Sim}(3)$. The resulting relative
% constraints are incorporated into the global factor graph to optimise the
% keyframe trajectory and dense reconstruction.

Keyframes are selected based on camera motion and scene covisibility: a new
keyframe is created when the estimated translation or rotation of the current frame relative to the
previous keyframe exceeds predefined thresholds, or when the number of
covisible features falls below a predefined threshold. Once the local factor graph contains a fixed number of keyframes, the global factor graph is triggered and updated.%The global graph isupdated after a fixed number of selected keyframes, \eg~five. 

\begin{table*}[t]
    \centering
    \caption{Relative pose error (RPE) on KITTI odd-numbered sequences and
    NCD Short. Results are reported over trajectory intervals of 5, 10, 50,
    and 100\,\si{\metre}. $t_{\mathrm{rel}}$ (\si{\metre}) and
    $r_{\mathrm{rel}}$ (\si{\degree}) denote translational and rotational RPE,
    respectively. Lower is better.}
    \label{tab:trajectory_results}

    % \scriptsize
    % \setlength{\tabcolsep}{1.3pt}
    \renewcommand{\arraystretch}{0.8}

    \resizebox{0.95\textwidth}{!}{%
    \begin{tabular}{l c cc cc cc cc cc cc}
        \toprule
        \multirow{2}{*}{\textbf{Method}} &
        \multirow{2}{*}{\textbf{Int.}} &
        \multicolumn{2}{c}{\textbf{KITTI 01}} &
        \multicolumn{2}{c}{\textbf{KITTI 03}} &
        \multicolumn{2}{c}{\textbf{KITTI 05}} &
        \multicolumn{2}{c}{\textbf{KITTI 07}} &
        \multicolumn{2}{c}{\textbf{KITTI 09}} &
        \multicolumn{2}{c}{\textbf{NCD Short}} \\
        \cmidrule(lr){3-4}
        \cmidrule(lr){5-6}
        \cmidrule(lr){7-8}
        \cmidrule(lr){9-10}
        \cmidrule(lr){11-12}
        \cmidrule(lr){13-14}

        & &
        $t_{\mathrm{rel}}$ & $r_{\mathrm{rel}}$ &
        $t_{\mathrm{rel}}$ & $r_{\mathrm{rel}}$ &
        $t_{\mathrm{rel}}$ & $r_{\mathrm{rel}}$ &
        $t_{\mathrm{rel}}$ & $r_{\mathrm{rel}}$ &
        $t_{\mathrm{rel}}$ & $r_{\mathrm{rel}}$ &
        $t_{\mathrm{rel}}$ & $r_{\mathrm{rel}}$ \\
        \midrule

        \multirow{4}{*}{ORB-SLAM3~\cite{9440682}}
        & $5$   & $\mathbf{0.15}$ & $\mathbf{0.05}$ & $\mathbf{0.08}$ & $\mathbf{0.07}$ & $\mathbf{0.08}$ & $\mathbf{0.21}$ & $\mathbf{0.10}$ & $\mathbf{0.14}$ & $\mathbf{0.09}$ & $\mathbf{0.08}$ & $0.81^*$ & $2.45^*$\\
        & $10$  & $\mathbf{0.31}$ & $\mathbf{0.08}$ & $\underline{0.15}$ & $\mathbf{0.10}$ & $\mathbf{0.12}$ & $\mathbf{0.24}$ & $\mathbf{0.14}$ & $\mathbf{0.20}$ & $\underline{0.17}$ & $\mathbf{0.11}$ &$1.62^*$ & $2.65^*$ \\
        & $50$  & $\mathbf{1.48}$ & $\underline{0.26}$ & $\underline{0.61}$ & $\mathbf{0.23}$ & $\mathbf{0.44}$ & $\underline{0.43}$ & $\mathbf{0.41}$ & $\mathbf{0.44}$ & $\underline{0.70}$ & $\mathbf{0.34}$ & $6.93^*$ & $3.24^*$ \\
        & $100$ & $\mathbf{2.69}$ & $\mathbf{0.35}$ & $\underline{1.11}$ & $\mathbf{0.30}$ & $\mathbf{0.79}$ & $\underline{0.59}$ & $\mathbf{0.69}$ & $\mathbf{0.60}$ & $\underline{1.23}$ & $\mathbf{0.48}$ &$10.90^*$ & $3.72^*$ \\
        \midrule

        \multirow{4}{*}{DROID-SLAM~\cite{NEURIPS2021_89fcd07f}}
        & $5$   & $5.28$ & $\mathbf{0.05}$ & $\mathbf{0.08}$ & $\underline{0.08}$ & $0.10$ & $\underline{0.22}$ & $0.40$ & $0.66$ & $\mathbf{0.09}$ & $\underline{0.11}$ & $\underline{2.65}$ & $\underline{2.80}$ \\
        & $10$  & $7.30$ & $\mathbf{0.08}$ & $\mathbf{0.14}$ & $\underline{0.11}$ & $\underline{0.14}$ & $\underline{0.27}$ & $0.86$ & $1.46$ & $\mathbf{0.15}$ & $\underline{0.17}$ & $\underline{5.05}$ & $\underline{3.08}$ \\
        & $50$  & $32.39$ & $\mathbf{0.25}$ & $\mathbf{0.53}$ & $\underline{0.25}$ & $\underline{0.45}$ & $\mathbf{0.38}$ & $2.32$ & $3.72$ & $\mathbf{0.58}$ & $\underline{0.39}$ & $\underline{21.05}$ & $\underline{5.31}$ \\
        & $100$ & $60.27$ & $\underline{0.36}$ & $\mathbf{0.99}$ & $\underline{0.36}$ & $\underline{0.80}$ & $\mathbf{0.51}$ & $5.64$ & $5.52$ & $\mathbf{1.03}$ & $\underline{0.52}$ & $\underline{36.50}$ & $\underline{7.47}$ \\
        \midrule

        \multirow{4}{*}{VGGT-SLAM 2.0~\cite{maggio2026vggt}}
        & $5$   & --- & --- & $1.30$ & $0.88$ & $3.06$ & $7.49$ & $1.58$ & $2.50$ & $3.38$ & $1.07$ & $16.27$ & $9.23$ \\
        & $10$  & $7.57$ & $0.31$ & $2.43$ & $1.17$ & $5.98$ & $9.26$ & $3.09$ & $2.92$ & $4.89$ & $1.67$ & $28.91$ & $16.56$ \\
        & $50$  & $33.01$ & $1.51$ & $11.02$ & $2.03$ & $28.77$ & $13.25$ & $19.78$ & $4.65$ & $22.78$ & $4.24$ & $86.88$ & $50.12$ \\
        & $100$ & $59.28$ & $3.14$ & $18.17$ & $2.79$ & $53.08$ & $17.49$ & $61.40$ & $4.45$ & $44.05$ & $5.19$ & $108.99$ & $60.48$ \\
        \midrule

        % Ours: committed kitti_config.yaml / ncd_config.yaml, full sequences, one seed, stereo landmark anchors, no lidar input;
        % NCD scored against the camera-frame reference (rooster_0_camRDF.txt). KITTI 01 fails (front-end degeneracy on the highway).
        % DROID-SLAM NCD Short rerun on our side (front end, no global BA, camera-frame reference): t_rel/r_rel 0.47/2.40, 0.93/2.50,
        % 4.18/2.93, 7.34/3.19 at 5/10/50/100 m. The row below (2.65 m at 5 m) looks scored against the LiDAR-frame poses; please confirm.
        % NOTE: Placeholder for now based on lidar guided NCD run of 8000 frames please replace the following values with the full experiment.
        \multirow{4}{*}{\textbf{Ours}}
        & $5$   & $\underline{0.35}$ & $\underline{0.23}$ & $\underline{0.10}$ & $0.09$ & $\underline{0.09}$ & $0.23$ & $\underline{0.16}$ & $\underline{0.30}$ & $\underline{0.12}$ & $0.12$ & $\mathbf{0.43}$ & $\mathbf{2.00}$ \\
        & $10$  & $\underline{0.94}$ & $\underline{0.24}$ & $0.17$ & $0.12$ & $0.15$ & $0.32$ & $\underline{0.22}$ & $\underline{0.43}$ & $0.23$ & $0.18$ & $\mathbf{0.63}$ & $\mathbf{2.38}$ \\
        & $50$  & $\underline{3.85}$ & $0.54$ & $0.64$ & $0.33$ & $0.54$ & $0.52$ & $\underline{0.57}$ & $\underline{0.64}$ & $0.98$ & $0.41$ & $\mathbf{2.14}$ & $\mathbf{3.91}$ \\
        & $100$ & $\underline{7.12}$ & $0.64$ & $1.21$ & $0.48$ & $1.03$ & $0.76$ & $\underline{1.08}$ & $\underline{0.90}$ & $1.82$ & $0.55$ & $\mathbf{4.19}$ & $\mathbf{5.12}$ \\

\bottomrule
% \vspace{1mm}

    \end{tabular}
    }
    \par\vspace{1mm}{\footnotesize Note1: Empty cells ($-$) indicate insufficient valid estimate--groundtruth pairs for metric computation. \\Note2: ORB-SLAM3 results (indicated by $^*$) on NCD are computed only up to frame 5050 (out of 15300), after which feature tracking failed. }%Empty cells indicate failure or divergence resulting in insufficient estimate--ground-truth pairs for metric computation.}
    % NOTE: threeparttable cannot wrap a resizebox (it produced LaTeX errors and "??" in the PDF), so the table note is set as a plain line with the same wording.
    \vspace{-4mm}
\end{table*}

We perform dense reconstruction using the multi-view configuration of
MapAnything on the selected keyframes in each inference window. We provide
the camera intrinsics and front-end pose estimates to the model, but do not use
raw stereo depth as a direct input. In our experiments, conditioning on raw
stereo depth distrupts the internal consistency of the predicted depth maps,
likely because stereo triangulation uncertainty increases approximately quadratically with distance.
Instead, we metrically calibrate the predicted geometry afterwards using the
optimised stereo landmarks from the corresponding local factor graph (\Cref{sec:metric_depth}).

\subsection{Datasets and Evaluation Metrics}
\label{sec:datasets_metrics}

\paragraph{KITTI Odometry}
We use the KITTI Odometry benchmark~\cite{geiger2012we} to evaluate large-scale trajectory estimation. KITTI provides calibrated stereo imagery together with reference trajectories over urban, residential, and highway environments. Its long sequences allow us to evaluate the scalability and long-term consistency of the proposed hierarchical factor-graph formulation. Due to space limitations, We report quantitative results on the odd-numbered sequences, while sequence 00 is used for qualitative analysis of the estimated trajectory and dense reconstruction.
% \milad{In addition to quantitative trajectory evaluation, we visualise the reconstructed dense metric maps to assess their consistency over large-scale motion.}

\paragraph{Newer College Dataset}
The Newer College Dataset (NCD)~\cite{ramezani2020newer} contains stereo imagery with a resolution of $848\times480$, captured using an Intel Realsense-D435, together with high-accuracy reference collected over diverse outdoor areas. 
% \sagun{We use the Short Experiment sequence, and every method, including ours, runs on the stereo calibration distributed with the dataset.}
Its varied outdoor structure provides a suitable setting for evaluating visual odometry performance and dense depth prediction for complex scene geometry. We use the Short Experiment run and process every third stereo frame, corresponding to an effective input rate of $10~\mathrm{Hz}$.
%We use NCD to evaluate both trajectory estimation and dense $3$D reconstruction under complex scene geometry. Its varied outdoor structure provides a suitable setting for assessing visual odometry performance as well as the spatial consistency and geometric quality of the reconstructed dense maps over extended trajectories. 
% \sagun{The RPE is computed against the reference trajectory expressed in the camera frame, since the reference poses given in the LiDAR frame leave the ATE unchanged but invalidate the RPE.}
% the DROID-SLAM NCD Short row is the shared VSLAM-LAB run scored against the camera-frame reference (checked 15 Sep: its groundtruth.csv equals rooster_0_camRDF.txt).

\paragraph{In-house Run}
Our in-house platform (shown in~\Cref{fig:hero}) is equipped with a ZED-X stereo camera with a $12\mathrm{cm}$ baseline, capturing $960\times600$ images at $15~\mathrm{Hz}$. A Livox LiDAR and its integrated IMU are used by an independent LiDAR--inertial SLAM system to obtain a reference trajectory. The resulting LiDAR reconstruction is used as %a prior LiDAR map serves as 
a geometric reference for evaluating dense-map accuracy. We selected a $\sim$ 40~\si{\metre} sequence whose start and end positions coincide, providing a controlled setting for assessing both local map accuracy and accumulated consistency over the full run. 

\paragraph{Evaluation Metrics}
Trajectory accuracy is evaluated using relative pose error (RPE) functions implemented in the \texttt{evo} package~\cite{grupp2017evo}. We report the root mean squared error (RMSE) of the translational ($t_{\mathrm{rel}}$) and rotational ($r_{\mathrm{rel}}$) components separately, over travelled-distance intervals of $5$, $10$, $50$, and $100$~\si{\metre}, taking all pose pairs separated by each interval rather than consecutive segments.
For quantitative dense-map evaluation, we use the in-house run and compare
the reconstructed map against the reference LiDAR map using the Chamfer distance, the mean of the two average
nearest-neighbour distances between the clouds, together with precision, recall and the F-score at a
$10$~\si{\centi\metre} threshold. Precision is the fraction of reconstructed points whose nearest reference point
lies within this distance, recall is the fraction of reference points whose nearest reconstructed point lies within
it, and the F-score is their harmonic mean. We score only points within $6$~\si{\metre} of the estimated
trajectory and voxel downsample the reference map to $5$~\si{\centi\metre}, while each reconstruction is scored
at its native density. Although this work focuses on odometry, we additionally report the absolute trajectory error (ATE) as RMSE after rigid alignment to the reference trajectory to provide an overall measure of accumulated error and deviation from ground truth.

\subsection{Baselines}
\label{sec:baselines}

Using VSLAM-LAB~\cite{11247218}, we compare against representative classical and learning-based visual SLAM systems. ORB-SLAM3~\cite{9440682} serves as a strong feature-based geometric baseline operating directly on calibrated stereo imagery. DROID-SLAM~\cite{NEURIPS2021_89fcd07f} represents learning-based visual SLAM with recurrent pose and depth refinement and differentiable dense bundle adjustment. VGGT-SLAM 2.0~\cite{maggio2026vggt} is included as a recent hybrid approach that combines feed-forward geometric prediction, using VGGT~\cite{wang2025vggt}, with global factor-graph optimisation. For evaluating dense-map consistency on the in-house run, we compare against ScaRF-SLAM~\cite{zhang2026scarf} and DROID-SLAM~\cite{NEURIPS2021_89fcd07f} as two separate baseline methods.

% Note that given the proposed approach does not contain loop closure, for fair comparison we have disabled loop closure detection and relocalisation in the baselines.
As our approach currently focuses on odometry and does not incorporate loop
closure, we disable loop-closure detection and relocalisation in the baselines
for a fair comparison.
For all baselines, we use the same image sequences and camera calibration whenever supported. Before evaluation, estimated trajectories are aligned to the reference frame using Umeyama alignment implemented in \texttt{evo}. For metric stereo methods, only rigid alignment is applied, without scale correction. Since VGGT-SLAM 2.0 operates on monocular input and does not inherently recover metric scale, we additionally allow a global scale correction during alignment.

\subsection{Trajectory Evaluation}
\label{sec:trajectory_results}

\Cref{tab:trajectory_results} reports the translational and rotational RPE on
the selected KITTI and NCD sequences. On KITTI, our method remains competitive with
ORB-SLAM3 and DROID-SLAM on sequences $03$, $05$, and $09$, while showing substantially
greater robustness than the learning-based baselines on the more challenging
sequences $01$ and $07$. In particular, at the $100\,\mathrm{m}$ interval on
sequence $01$, our method achieves a translational RPE of $7.12\,\mathrm{m}$,
compared with $60.27\,\mathrm{m}$ for DROID-SLAM and $59.28\,\mathrm{m}$ for
VGGT-SLAM 2.0.

The difference becomes more pronounced on NCD, where the learning-based
baselines degrade considerably under the more complex motion and scene
geometry. Our method maintains substantially lower translational and rotational errors across all evaluated intervals, while ORB-SLAM3 fails to complete the full sequence, tracking only $480$ of $1410.5$~\si{m} due to feature-tracking failure. We hypothesise that this is caused by the substantial illumination variation in NCD, particularly in the garden section, together with the weaker stereo geometry provided by the Intel RealSense D435. Accordingly, the ORB-SLAM3 results reported in ~\Cref{tab:trajectory_results} are computed only up to the failure point. These results suggest that retaining model-based stereo odometry while exploiting learned geometry at the submap level provides greater resilience when either feature-based or learning-based SLAM becomes unreliable.

\begin{figure}[t]
    \centering
    \includegraphics[width=0.99\linewidth]{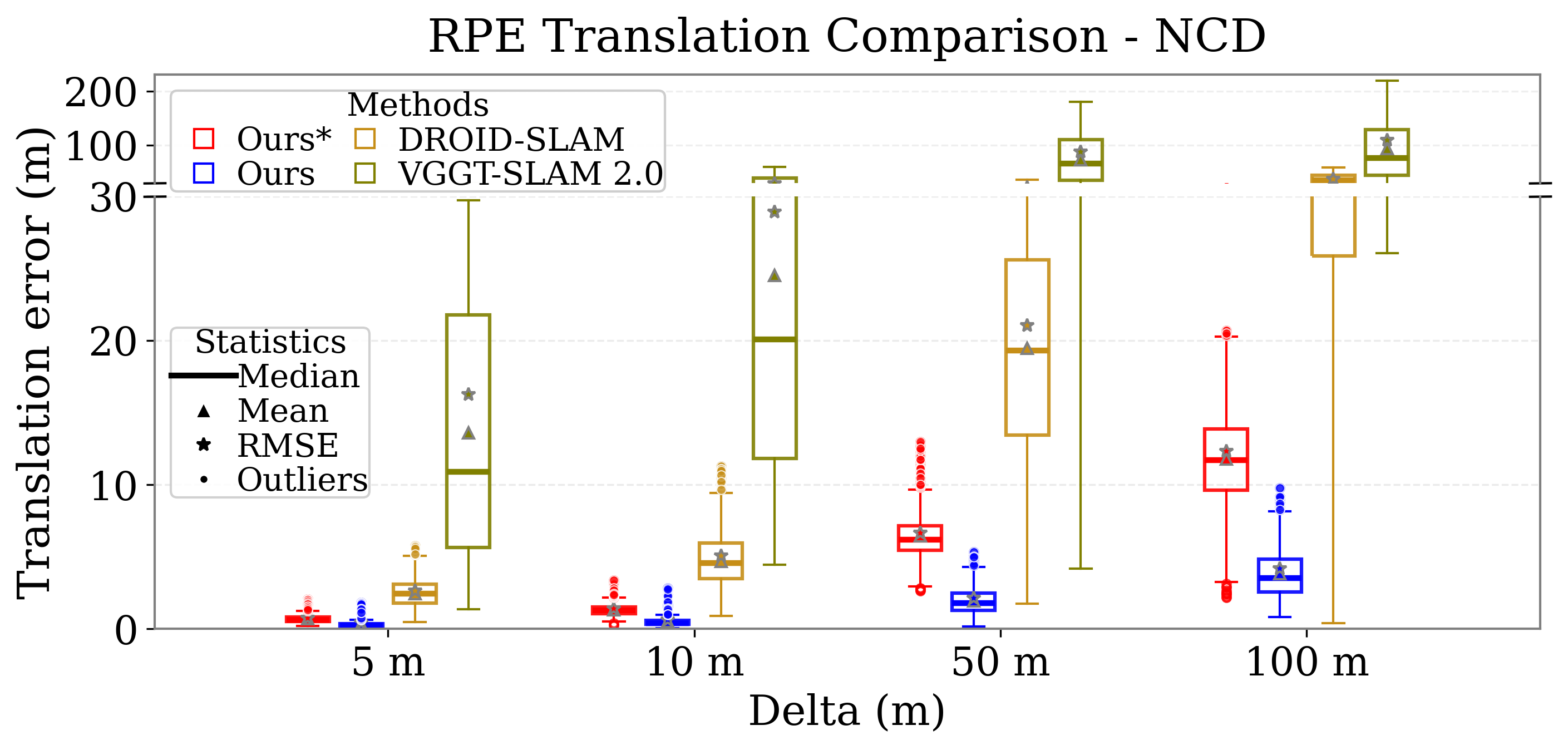}
    % \vspace{1mm}
    \includegraphics[width=0.99\linewidth]{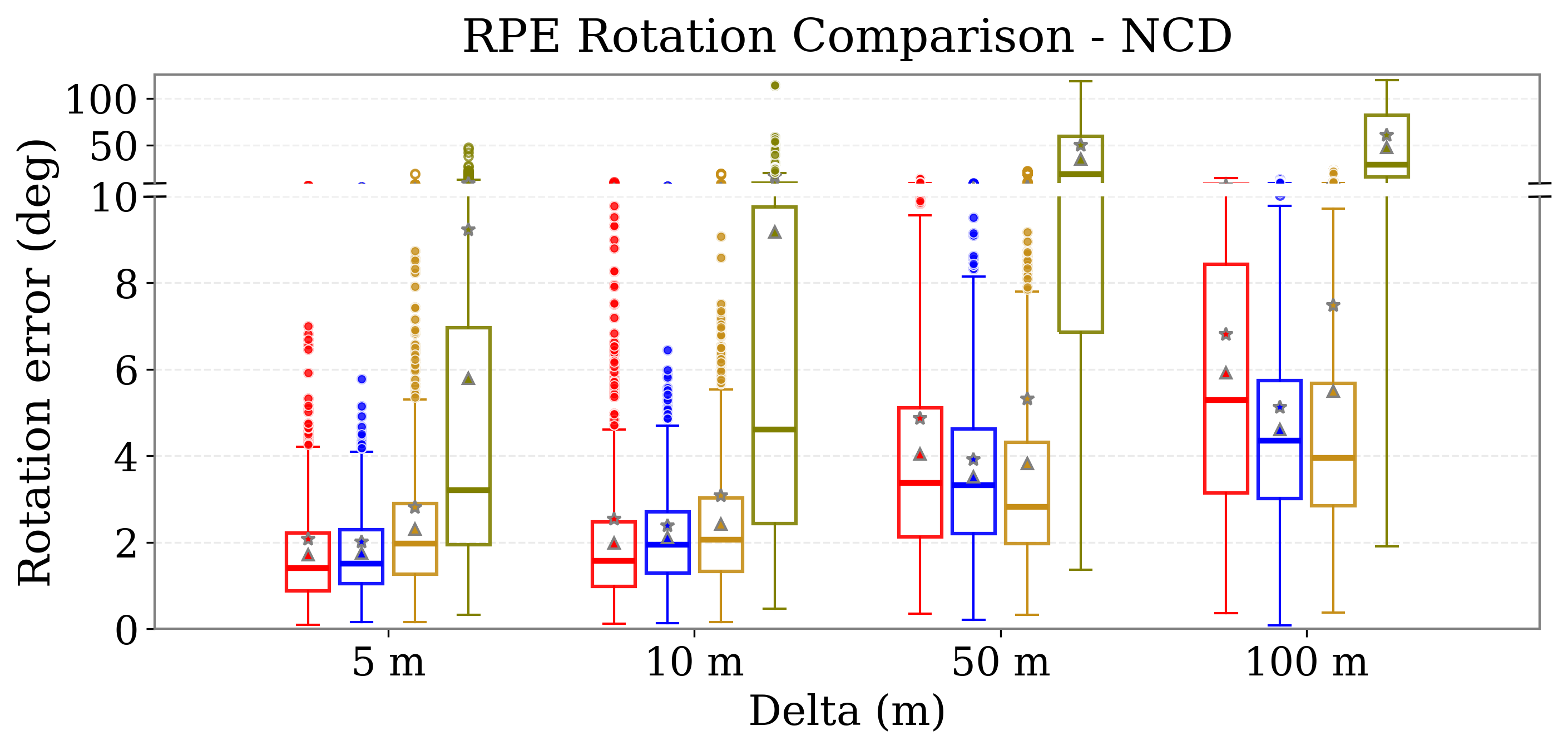}
    \caption{Statistical analysis of translational (top) and rotational (bottom) RPE on the NCD Short Experiment. $^*$ indicates the stereo visual odometry alone with the depth model disabled.}
    \label{fig:ncd_box_plots}
    \vspace{-4mm}
\end{figure}

\begin{figure}[t]
    \centering
    \includegraphics[width=0.99\linewidth]{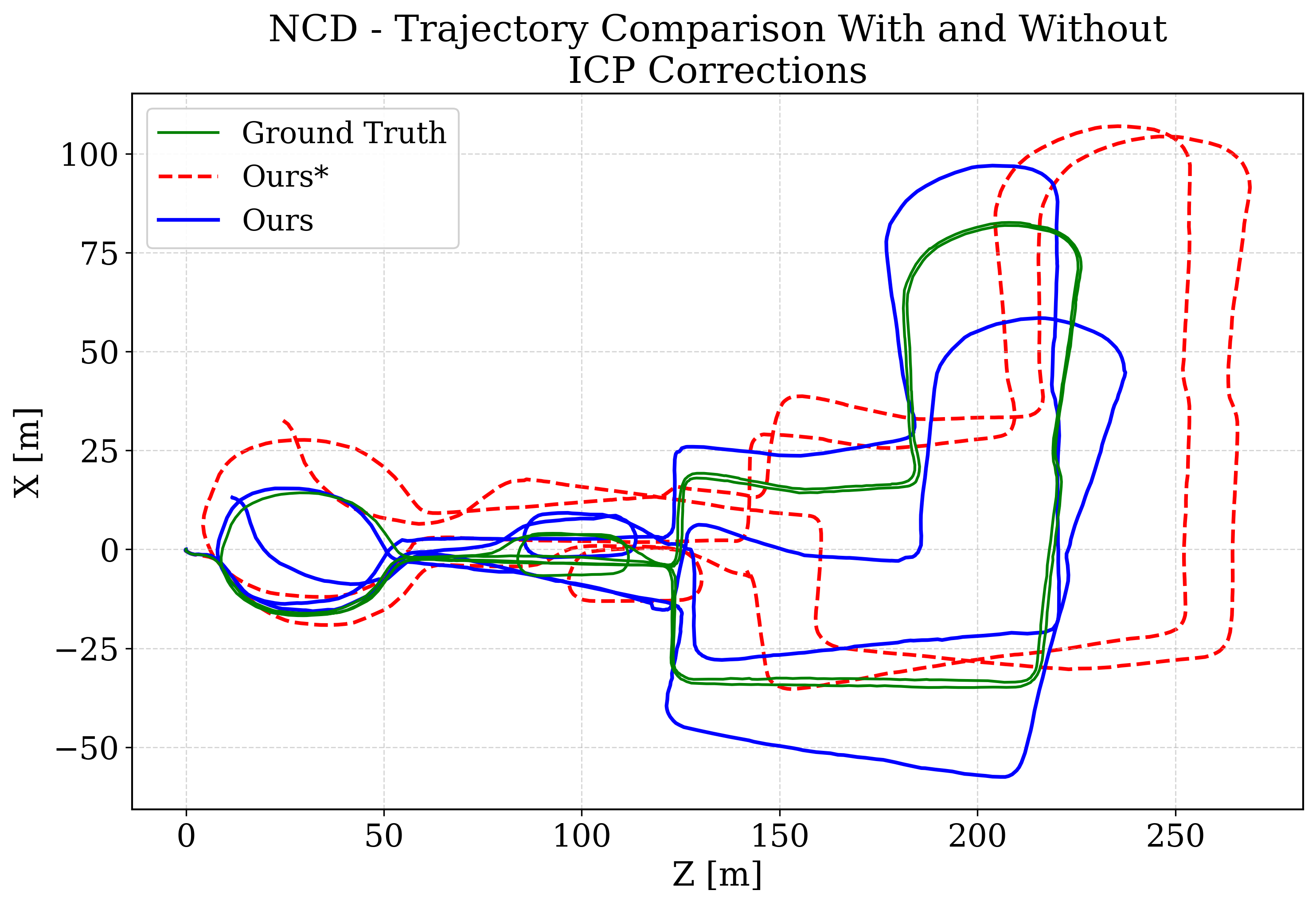}
    \caption{Comparison of the estimated trajectories on the NCD Short Experiment showing improved long-term trajectory consistency with dense-map constraints.~$^*$ indicates the stereo visual odometry alone with the depth model disabled.}
    \label{fig:dsvo_vs_svo}
    \vspace{-2mm}
\end{figure}

% A key objective of our design is not to replace the geometric estimator with learned predictions, but to use the two representations complementarily. Consequently, when dense prediction or registration becomes unreliable, the local SVO continues to provide metric motion estimates, while reliable dense constraints improve the globally optimised trajectory.
Additionally, \Cref{fig:ncd_box_plots} compares the RPE distributions of our full method against the learning-based baselines and our model-based SVO variant without dense-map constraints (denoted by Ours$^*$). Incorporating dense-map registration substantially reduces translational drift, particularly over longer intervals, demonstrating the effectiveness of metric depth calibration and the benefit of integrating the resulting metric dense maps into the VO framework through intra- and inter-submap alignment. Overall, the ICP-based constraints improve odometry consistency, though their benefit depends on the quality of the dense geometry used for alignment.
\Cref{fig:dsvo_vs_svo} further compares the estimated trajectories of our full method and Ours$^*$, showing improved long-term trajectory consistency when dense-map constraints are incorporated.

\subsection{Dense Map Evaluation}
\label{sec:mapping_results}

\begin{figure*}[t]
    \centering
     \includegraphics[width=1.0\textwidth, trim=0 4cm 0 0, clip]{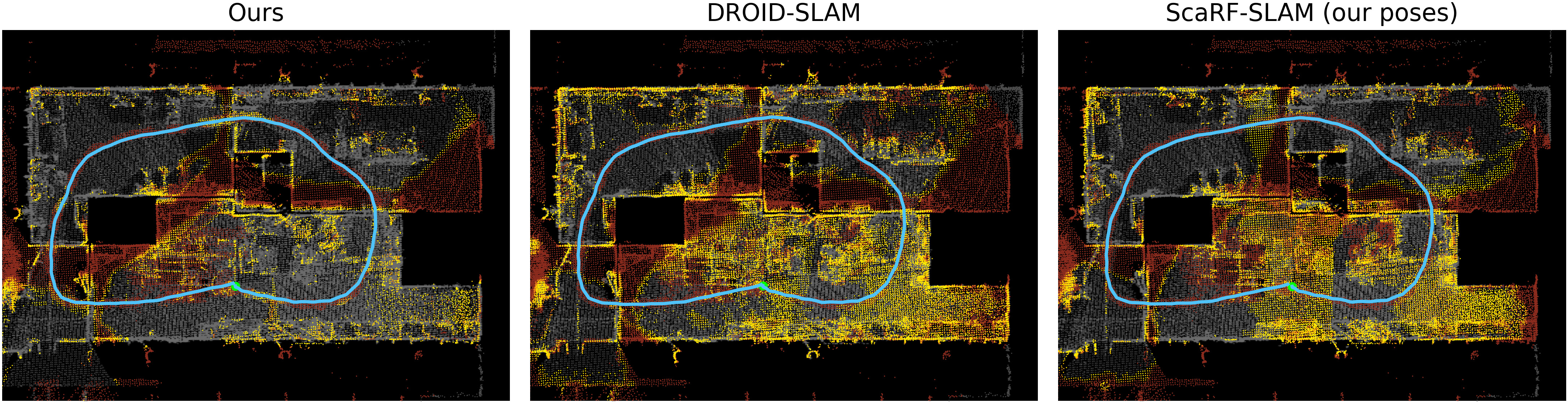}
    \caption{Dense-map comparison on the in-house sequence: Ours (left), DROID-SLAM (middle), and ScaRF-SLAM (right). ScaRF-SLAM uses the poses estimated by our stereo VO, while DROID-SLAM uses its own front-end estimates; the estimated trajectory of each method are shown in blue. The reference LiDAR map is coloured according to its distance to each reconstruction: grey denotes surfaces within $10\,\mathrm{cm}$, while deviations increase from yellow at $10\,\mathrm{cm}$ to red beyond $30\,\mathrm{cm}$. Regions without reconstructed surfaces therefore remain red, making the visualisation largely insensitive to reconstruction density. DROID-SLAM retains depth only for well-matched keyframes and consequently does not produce a comparably dense map.}
    % NOTE: coverage direction, native density (--row2 coverage); recommended
    \label{fig:inhouse_panel}
\end{figure*}

\begin{figure*}[t]
    \centering

    \begin{subfigure}[t]{0.32\textwidth}
        \centering
        \includegraphics[width=\linewidth, trim=0 1cm 0 1cm, clip]{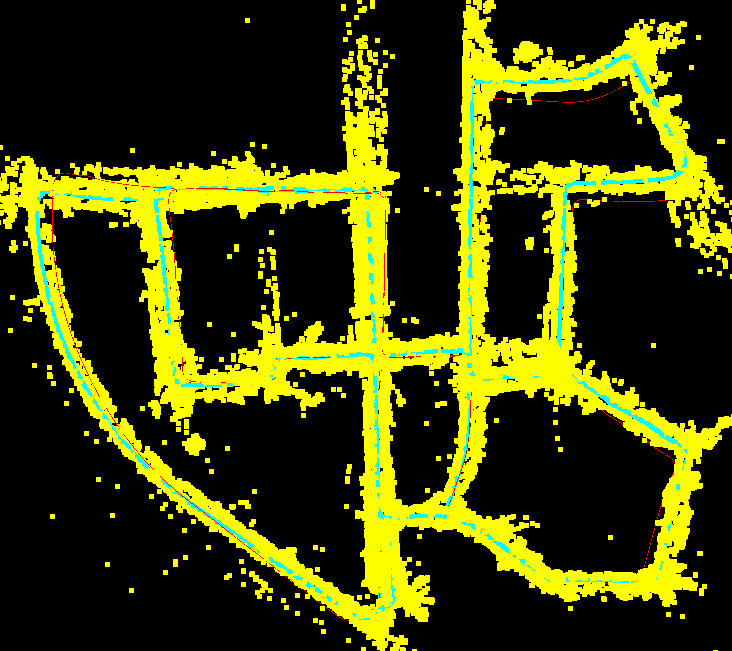}
        % \caption{}
        % \label{fig:sub1}
    \end{subfigure}
    \hfill
    \begin{subfigure}[t]{0.32\textwidth}
        \centering
        \includegraphics[width=\linewidth, trim=0 1cm 0 1cm, clip]{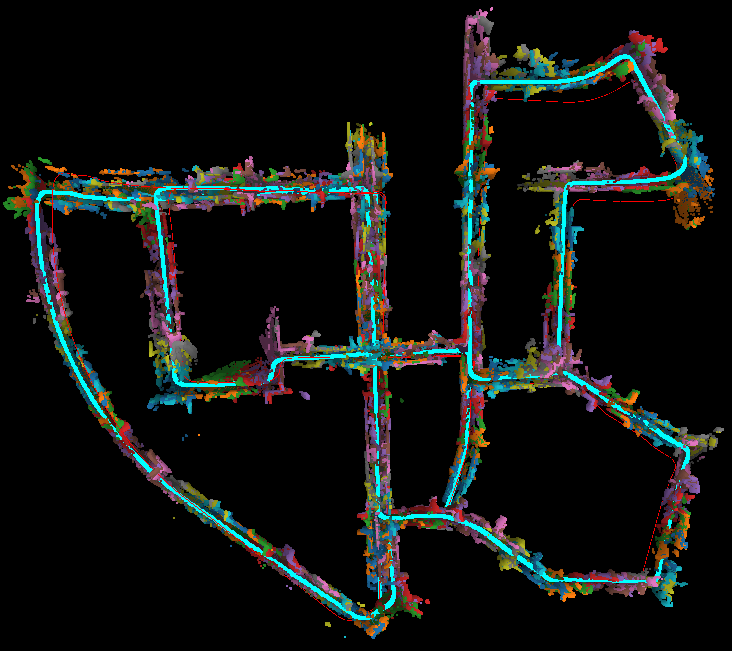}
        % \caption{}
        % \label{fig:sub2}
    \end{subfigure}
    \hfill
    \begin{subfigure}[t]{0.32\textwidth}
        \centering
        \includegraphics[width=\linewidth, trim=0 1cm 0 1cm, clip]{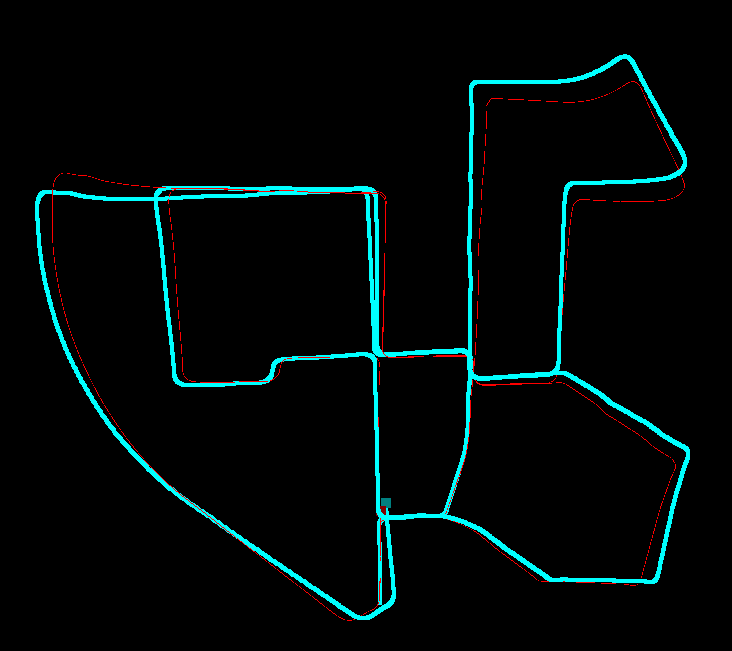}
        % \caption{}
        % \label{fig:kitti_00}
    \end{subfigure}

    \caption{Qualitative results on KITTI sequence $00$. Left: sparse metric landmarks estimated from local factor graphs. Middle: consecutively registered dense submaps. Right: estimated trajectory (cyan) and ground truth (red).}
    % \label{fig:three_subplots}
    \label{fig:kitti_00}
    \vspace{-5mm}
\end{figure*}

\begin{table}[t]
    \centering
    \caption{Dense-map accuracy and ATE on the in-house run, all systems without loop closure. F-score, precision and recall at $10\,\mathrm{cm}$ in \%, ATE and Chamfer distance in \si{\metre}. All map scores follow one rigid alignment of the estimated trajectory to the reference. }
    \label{tab:inhouse_map}
    \setlength{\tabcolsep}{4pt}
    \resizebox{\columnwidth}{!}{%
        \begin{tabular}{lccccc}
            \toprule
            \textbf{Method} & \textbf{ATE $\downarrow$} & \textbf{F-score $\uparrow$} & \textbf{Precision $\uparrow$} & \textbf{Recall $\uparrow$} & \textbf{Chamfer $\downarrow$} \\
            \midrule
            % Precision read from the dense_map_eval fscore line of each run: ScaRF ORB 20260905_1930/metrics_orb, ScaRF our-VO 20260909_ourvo/metrics_orb, DROID 20260908_1635/metrics_odo, ours 20260914_w81_y_rr_base_s3/metrics.
            \makecell[l]{ScaRF-SLAM\\(ORB-SLAM3 poses)} & $0.074$ & $55.2$ & $70.6$ & $45.3$ & $0.192$ \\
            \makecell[l]{ScaRF-SLAM\\(Our VO poses)} & $0.053^{\dagger}$ & $54.1$ & $71.9$ & $43.4$ & $0.193$ \\
            \makecell[l]{DROID-SLAM\\(Front end)} & $0.055$ & $54.6$ & $\mathbf{90.2}$ & $39.1$ & $0.179$ \\
            \textbf{Ours} & $\mathbf{0.053}$ & $\mathbf{64.9}$ & $68.2$ & $\mathbf{61.9}$ & $\mathbf{0.162}$ \\
            \bottomrule
        \end{tabular}

        }
        \par\vspace{1mm}
        {\footnotesize\raggedright
        $^\dagger$ScaRF-SLAM keeps input poses fixed and optimises only depth scale; thus, with our VO poses, its ATE is identical to ours.\par
}
    \vspace{-5mm}
\end{table}

To evaluate the dense reconstruction quality, we compare maps reconstructed by our method, DROID-SLAM, and ScaRF-SLAM on the in-house sequence against the reference LiDAR map, with results reported from a single run per method. As shown in~\Cref{tab:inhouse_map}, our method achieves the highest F-score ($64.9\%$ versus $54.1$--$55.2\%$) and recall
($61.9\%$ versus $39.1$--$45.3\%$), together with the lowest Chamfer distance
and ATE. The improvement is driven primarily by recall rather than precision: DROID-SLAM retains depth only for well-matched keyframes, yielding the highest precision ($90.2\%$ versus $68.2\%$ for ours) but the lowest scene coverage, while ScaRF-SLAM filters low-confidence depth during fusion, also limiting coverage. This trend is also visible in~\Cref{fig:inhouse_panel}, where a larger portion of the reference LiDAR map remains uncovered or misaligned for the baselines. 

In contrast, our metrically calibrated dense predictions recover a larger fraction of the observed scene, providing a denser geometric representation that is potentially more useful for downstream tasks such as object detection and semantic segmentation. As we extend the framework towards a complete SLAM system, we will exploit fusion strategies to improve precision while retaining this high scene coverage.

% This trend is also visible in~\Cref{fig:inhouse_panel}. The reference LiDAR map
% remains grey over a larger portion of the environment for our reconstruction,
% indicating agreement within $10\,\mathrm{cm}$, whereas larger regions remain
% uncovered or misaligned for the baselines. These results show the benefit of
% combining sparse metric geometry from the local factor graphs with dense GFM
% predictions and subsequent global alignment.

We additionally provide a qualitative large-scale reconstruction on KITTI
sequence $00$ in~\Cref{fig:kitti_00}. Leveraging locally optimised sparse landmarks, the metrically calibrated keyframe depth maps are accumulated within local submaps, producing a dense reconstruction that remains spatially consistent with the estimated trajectory over the sequence.

\begin{table}[t]
    \centering
    \caption{Average runtime per frame on the in-house run.}
    \label{tab:runtime}
    \resizebox{0.90\columnwidth}{!}{%
        \begin{tabular}{lrr}
            \toprule
            \textbf{Component} & \textbf{Laptop} & \textbf{AGX Orin} \\
            & [ms/frame] & [ms/frame] \\
            \midrule
            Front-end (feature tracking)    & $16.9$ & $32.0$ \\
            Local FGO                       & $48.9$ & $72.4$ \\
            GFM inference                   & $10.6$ & $23.8$ \\
            Metric depth alignment          & $2.1$  & $5.4$ \\
            Back-end (GICP + global FGO)    & $15.7$ & $29.1$ \\
            \midrule
            \textbf{Total}                  & $94.2$ & $162.7$ \\
            % Wall clock, not reported in the table (the table is the sum of the explicit stages, as before):
            %   laptop  110.9 ms/frame  (175.28 s over 1581 frames)   -> 15% above the stage sum
            %   Orin    196.4 ms/frame  (310.55 s over 1581 frames)   -> 21% above the stage sum
            % The gap is work inside no timed stage: image loading, PnP RANSAC, process start.
            % Source: <run>/wall_clock.txt and the pipeline's own summary line in <run>/logs/run.log
            %   laptop 20260916_1424_hl_timing, Orin 20260917_0016_orin_hl_timing.
            \bottomrule
        \end{tabular}%
    }
    \vspace{-5mm}
\end{table}

\subsection{Computational Analysis}
\label{sec:runtime}
We analyse the computational cost of the main system components, including feature tracking, local factor-graph optimisation, GFM inference, metric depth alignment, dense registration, and global optimisation. Since dense inference and registration are performed only on selected keyframes, their computational cost is distributed over the input sequence rather than incurred at every frame.
Runtime is evaluated over $1581$ frames on a laptop equipped with an Intel Core Ultra 9 285HX CPU and an NVIDIA RTX PRO 3000 Blackwell Laptop GPU with $12~\mathrm{GB}$ of memory, and on an NVIDIA Jetson AGX Orin at MAXN with pinned clocks. Averaged over the full sequence, the total processing time is about $94~\mathrm{ms}$ per frame on the laptop and $163~\mathrm{ms}$ ($1.7\times$) on the Orin. As shown in~\Cref{tab:runtime}, feature tracking and local factor-graph optimisation together account for $\sim70\%$ and $64\%$ of the runtime on the laptop and on the Orin, respectively, while dense inference contributes only $11\%$ and $15\%$ , because it is restricted to keyframes within submaps. This indicates that the main computational cost lies in visual tracking and incremental local optimisation rather than GFM inference, demonstrating the benefit of keeping dense prediction off the per-frame processing path. Future work will investigate asynchronous execution, GPU-based backend optimisation, and model acceleration to improve deployment efficiency.

\section{Conclusion}
\label{sec:conclusion}

We presented a stereo visual odometry framework that integrates geometric feed-forward depth prediction with a hierarchical factor-graph architecture for accurate localisation and dense metric mapping. At the local level, stereo and temporal multi-view observations provide metrically scaled camera poses, and $3$D landmarks. These optimised landmarks are then used to estimate a scale-shift transformation that metrically correct dense depth predictions from a geometric feed-forward model. At the global level, dense keyframe reconstructions are registered using GICP, while consecutive submaps are aligned in $\mathrm{Sim}(3)$ to correct residual scale inconsistencies. These geometric constraints are incorporated into a global factor graph, enabling long-term trajectory consistency and dense metric reconstruction while limiting expensive dense processing to selected keyframes.

Experiments on KITTI, Newer College, and in-house runs demonstrate the potential of combining sparse but reliable geometric estimation with dense learned scene reconstruction. The proposed formulation retains the metric consistency and robustness of model-based stereo visual odometry while exploiting feed-forward models to obtain substantially richer scene geometry. Our current system focuses on odometry and does not yet incorporate loop closure. Future work will therefore investigate loop-closure constraints for improved long-term global consistency, as well as computational optimisation and asynchronous processing toward real-time deployment.

% \section{Ideas Welcome}
% \begin{enumerate}
%   \item Obtain dense depth maps with stereo guidance for every $n$-th frame. May be used to provide initial estimates for $3$D landmarks (This would be overkill?).
%   \item Exploit these dense depth maps for 3D reconstruction.
%   \item Use the resulting dense global maps for local-to-global registration to introduce factors similar to those used in the VLI-SLAM framework.
%   \item Additionally, perform scan-to-scan registration between consecutive local dense depth maps to obtain an initial relative pose estimate (as a between-factor, or compose it with the previous global pose to initialize the current global pose.
%   \item May be get local maps from each $n$ frames. Then for each submap obtained from local optimization, use scan-scan registration to get a constraint between them.(sounds similar to above)
%   \item filter dynamic objects in tracker using deep net predictions?see how Gaussian-lic is doing
%   \item semantic tracker
% \end{enumerate}


% \section*{ACKNOWLEDGMENT}
% Include AFTER acceptance (DO NOT SUBMIT FOR REVIEW VERSION)

\balance{}

\bibliographystyle{IEEEtran} % use IEEEtran.bst style
\bibliography{IEEEabrv,ref}

\end{document}